# SI-FACT: Mitigating Knowledge Conflict via Self-Improving Faithfulness-Aware Contrastive Tuning

Shengqiang Fu

## Abstract

Large Language Models (LLMs) often generate unfaithful responses in knowledge-intensive tasks due to knowledge conflict—that is, a preference for relying on internal parametric knowledge rather than the provided context. To address this issue, we propose a novel self-improving framework, Self-Improving Faithfulness-Aware Contrastive Tuning (SI-FACT). The framework uses a self-instruct mechanism that allows the base LLM to automatically generate high-quality, structured contrastive learning data, including anchor samples, semantically equivalent positive samples, and negative samples simulating unfaithful scenarios. This approach significantly reduces the cost of manual annotation. Subsequently, contrastive learning is applied to train the model, enabling it to pull faithful responses closer and push unfaithful responses farther apart in the representation space. Experiments on knowledge conflict evaluation benchmarks ECARE_KRE and COSE_KRE show that the SI-FACT model based on Llama3-8B-Instruct improves the Contextual Recall Rate (CRR) by 6.2% over the best baseline method, while significantly reducing dependence on internal memory. The results indicate that SI-FACT provides strong effectiveness and high data efficiency in enhancing the contextual faithfulness of LLMs, offering a practical pathway toward building more proactive and trustworthy language models.

**Keywords:** Knowledge conflict, Contrastive learning, Self-instruct, Large language models, Contextual faithfulness

## 0. Introduction

Large Language Models (LLMs) have become a core driver for knowledge-intensive tasks, but their application in high-reliability scenarios, such as financial decision-making and medical diagnosis, faces a fundamental challenge: Unfaithful Generation[1]. This is particularly critical in fields like finance, where if the model generates content that contradicts the context or is unsubstantiated (i.e., hallucination[2]), it can lead to severe consequences[3, 4]. The core of this problem lies in knowledge conflict, especially Context-Memory Conflict—that is, when the internal knowledge (memory) solidified during the model's pre-training contradicts the immediately input context, LLMs often exhibit a systematic preference for internal memory, a form of stubbornness[5, 6]. This not only reveals a lack of dynamic adaptation to external knowledge but also poses a serious threat to their reliability in open environments.

Existing mitigation strategies, while having made some progress, still show significant limitations. Inference-time intervention methods (e.g., prompt engineering[7], neuron editing[8]) are lightweight and flexible, capable of temporarily alleviating some conflicts without modifying model parameters. However, these methods rely more on manual design or local operations, only superficially guiding the model to focus on the context, and struggle to address the model's inherent knowledge biases fundamentally[9]. Another approach is traditional supervised fine-tuning, which retrains the model on human-annotated data and can partially correct unfaithful generation. But this method is not only costly[10] but can also make the model overly dependent on external supervision signals, leading to overfitting and even catastrophic forgetting[11]. Therefore, neither lightweight inference-time interventions nor expensive supervised fine-tuning can fundamentally resolve the knowledge conflict problem.

To address these issues, this paper proposes a novel training framework called Self-Improving Faithfulness-Aware Contrastive Tuning (SI-FACT). We no longer view the model as an object requiring external correction but reshape it into an active learner capable of Self-Improvement. The core idea of SI-FACT is to let the model autonomously generate data to enhance contextual faithfulness for guiding its own optimization. This self-improvement paradigm, inspired by self-supervised learning[12], guides the LLM to actively distill and apply its existing internal knowledge to create a tailored training dataset for enhancing contextual faithfulness. The training data includes faithful answers

(positive samples) and various carefully designed unfaithful answers (negative samples). Subsequently, through contrastive learning, the model learns from its self-generated data, internalizing the principles for distinguishing between faithful and unfaithful answers.

This paradigm achieves a shift in the model's role from a passive knowledge recipient to an active self-adaptive optimizer. This paradigm shift from passive instruction to active learning not only alleviates the data bottleneck efficiently and at low cost but also fundamentally enhances the model's ability to adhere to the context when facing knowledge conflicts.

The main contributions of this paper are as follows:

(1) We propose the SI-FACT framework, pioneering a self-instructed contrastive data generation pipeline that enables active self-improvement of LLMs.

(2) Through extensive experiments on two challenging knowledge conflict benchmarks, we demonstrate the superiority of the SI-FACT framework over various baseline methods, improving model contextual faithfulness and revealing its high data efficiency.

(3) We provide a scalable path for building trustworthy LLMs, proving the potential of the "model self-improving" paradigm for optimizing key capabilities.

# 1. Related Work

This section reviews research related to mitigating knowledge conflicts in LLMs. Based on the timing of the intervention, existing methods can be broadly categorized into pre-hoc strategies applied during the training phase and post-hoc strategies[5] applied after training during inference. The SI-FACT framework proposed in this study belongs to a data-efficient pre-hoc strategy.

## 1.1 The Challenge of Knowledge Conflict in LLMs

Knowledge conflict is a core challenge for LLMs when integrating information from different sources. Depending on the source of the conflict, it can be categorized into context-memory conflict, inter-context conflict, and intra-memory conflict[5]. This paper focuses on context-memory conflict, which is the contradiction between the model's internal parametric knowledge and the external input context, and is one of the primary causes of unfaithful generation. To systematically evaluate model behavior in such scenarios, researchers have developed specialized benchmark datasets. For example, the ECARE_KRE and COSE_KRE datasets[8] used in this paper test whether a model adheres to its memory or follows the current context by providing machine-generated, misleading context that contradicts the model's internal knowledge.

## 1.2 Pre-hoc Strategies

Retrieval-Augmented Generation (RAG)[13] is a common method for mitigating knowledge obsolescence and hallucination by retrieving relevant documents from external knowledge bases before generation, providing the model with immediate and accurate information. However, RAG itself cannot fully resolve knowledge conflicts because the model may still ignore retrieved context that contradicts its internal knowledge[14]. Therefore, methods specifically designed to train the model to trust and remain faithful to the provided context are important complements to the RAG framework.

Continual Learning[15] and Model Editing[16] focus on incrementally updating or correcting knowledge within the model without causing catastrophic forgetting. These methods primarily address "how to learn new knowledge", whereas the focus of this paper is on "how to prioritize believing new information when facing conflict", which is a related but distinct problem.

## 1.3 Post-hoc Strategies

Post-hoc strategies do not require retraining the model and have garnered widespread attention due to their flexibility and low computational cost.

### 1.3.1 Prompt Engineering

The most direct intervention method is prompt engineering[17]. Simply adding instructions like "based on the context" to the prompt or structuring the input can guide the model to focus on the context. A more creative strategy is the Opin[7] method, which transforms factual questions into opinion-seeking questions (e.g., Context: Bob says "The capital of France is Paris."; Question: "According to Bob's view, what is the capital of France?"). This method assumes that when answering opinion-based questions, the model will more naturally rely on the provided text rather than stating its internal memory.

### 1.3.2 Decoding Strategies and Model Intervention

Context-aware Decoding (CAD)[18] is a more advanced inference-time technique. Its core idea is to amplify the probability of tokens that have a higher probability under the with context condition compared to the without context condition when generating each token. It effectively suppresses the model's tendency to generate answers based solely on internal knowledge by contrasting the output distributions under these two conditions, performing particularly well in knowledge conflict scenarios. Identifying and Reweighting Context-Aware Neurons (IRCAN)[8] is a more fine-grained intervention method. It aims to identify and enhance specific neurons crucial for processing contextual information. This method uses integrated gradients[19] to calculate a context-awareness contribution score for each neuron in the feedforward networks (FFNs) and then amplifies the weights of those context-aware neurons. In this way, IRCAN guides the model to pay more attention to contextual information during generation, thereby mitigating knowledge conflict.

# 2. Proposed Method

We utilize a self-improving paradigm in conjunction with contrastive learning to enhance the model's specific capability of contextual faithfulness. The SI-FACT framework aligns with the spirit of Self-Supervised Learning, with its core advantage being the automation of supervision signal generation. However, unlike classical self-supervised learning, our method does not passively extract signals from the inherent structure of the data. Instead, it actively leverages the model's existing knowledge to generate supervision signals specifically oriented toward contextual faithfulness. This makes acquiring high-quality contrastive data significantly easier, and through contrastive learning, the model effectively learns contextual faithfulness.

## 2.1 Framework Overview

The essence of the SI-FACT framework is a self-improvement loop, as shown in Fig. 1. In this loop, the LLM plays a dual role: it is both the teacher generating training data and the student learning from this data. This process aims to transform the abstract concept of faithfulness to context into a concrete signal that the model can learn and optimize in the representation space.

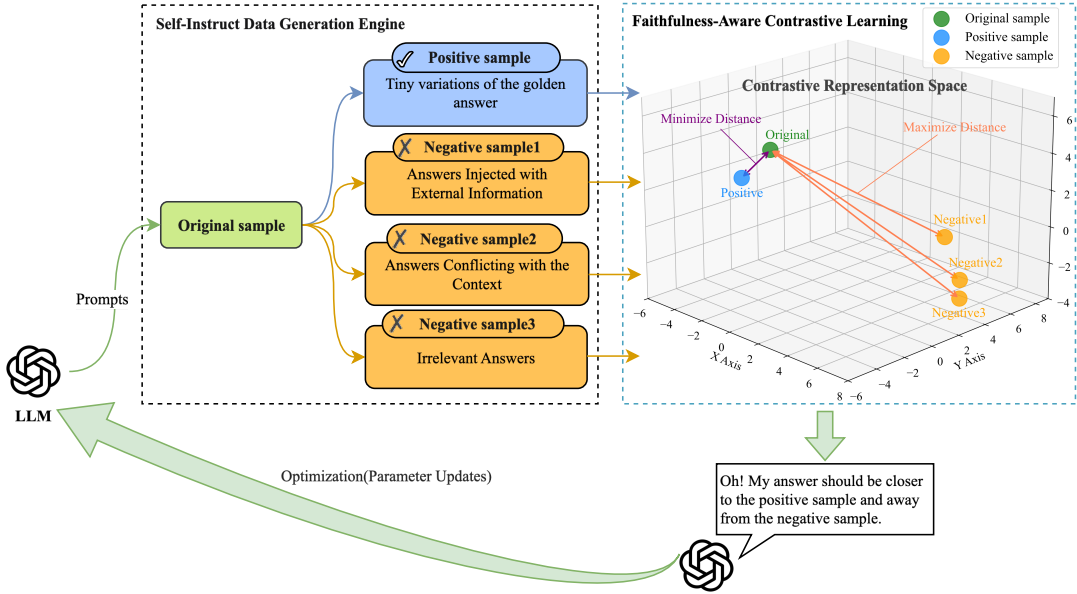

Fig 1 The framework of SI-FACT

This loop consists of the following key steps:

(1) Anchor Selection: Extract raw data triplet $(Context, Question, Answer_{golden})$ from a standard QA dataset (SQuAD)[20] to serve as anchor for learning.

(2) Self-Instruct Data Generation: Using the base LLM as the teacher, we design prompts to automatically produce a contrastive set for each anchor.

(3) Contrastive Learning: Transform the LLM into the student and train it using the generated contrastive dataset. This step forces the model to learn a representation space where the representations of faithful answers are pulled closer together and those of unfaithful answers are pushed apart. The distance between representations is measured using cosine similarity:

$$sim(x, y) = \frac{(x^T y)}{||x||\,||y||} \quad (1)$$

(4) Capability Internalization: After training, the model internalizes the capability of contextual faithfulness, completing one cycle of self-improvement. In principle, the improved model can generate higher-quality data, forming a virtuous cycle.

## 2.2 Self-Instruct Data Generation Engine

The driving force of the SI-FACT framework is its self-instruct data generation engine. This engine transforms the traditionally labor-intensive and expertise-dependent annotation process into a LLM-driven automated pipeline.

**Anchor and Positive Samples**

For each $(C, Q, A_{golden})$ triplet extracted from the source dataset, we use it as an anchor. To allow the model to learn semantic robustness towards contextually faithful answers, we construct a positive sample $A_{pos}$. Specifically, we provide the LLM with a prompt requiring it to rewrite the golden answer in different words while keeping the factual information completely unchanged.

**Negative Samples**

The design of negative sample is the most critical step in data generation. Instead of generating negative samples randomly, we design a set of structured instructions for the model, covering the three most critical types of unfaithfulness, forcing the model to learn fine-grained discriminative ability.

(1) Type 1: Answers Injected with External Information: These negative samples simulate the model hallucinating external information that not mentioned in the context.

(2) Type 2: Answers Conflicting with the Context: These negative samples deliberately alter or deny key information in the context, directly conflicting with the contextual content.

(3) Type 3: Irrelevant Answers: These negative samples might be based on some information in the context but do not directly and accurately answer the user's core question.

To illustrate this process concretely, Table 1 shows a complete contrastive learning sample.

Table 1　Example of Contrastive Learning Sample

| Field | Content | Description |
| --- | --- | --- |
| **Context** | Schwarzenegger was so enamored by the Humvee that he lobbied AM General to produce a civilian Humvee, which they did in 1992. He purchased the first two. | Provides basic facts about Schwarzenegger and the Humvee. |
| **Question** | In what year did AM General grant Schwarzenegger's wish for a street-legal Humvee? | Targets specific information in the context. |
| **Golden Answer** | In 1992. | Accurate answer extracted directly from the context. |
| **Positive Sample** | AM General launched the road-legal Humvee in 1992. | Tiny variations of the golden answer |
| **Negative Sample 1** | 1992, after Schwarzenegger promised a cameo in Terminator 3. | Adds information not mentioned in the context. |
| **Negative Sample 2** | In 1989, AM General finally relented. | Provides a wrong year (1989) conflicting with the context. |
| **Negative Sample 3** | Schwarzenegger famously owned two of the first civilian Hummers ever produced by AM General. | The answer, while based on the context, does not directly answer the "in what year" question. |

By systematically generating these three types of hard negative samples and one positive sample for each anchor, we provide the model with a rich, targeted learning environment that forces it to learn fine-grained contextual faithfulness discrimination capabilities.

### 2.3 Faithfulness-Aware Contrastive Learning

In the student phase, the model's goal is to learn the concept of contextual faithfulness as a core dimension of its representation space through contrastive learning.

**Representation Extraction**

For input triplet $(Context, Question, Answer)$, we concatenate them into a complete sequence and feed it into the LLM. We adopt a standard representation extraction method[21]: take the hidden state of the last token of the final transformer layer's output as the overall representation vector $h_\theta(C, Q, A)$ for this input sequence. This vector is considered to encode the model's comprehensive understanding of the entire input sequence, including its judgment on the answer quality.

**Optimization Objective and Loss Function**

Our optimization objective is to make the vectors representing faithful answers (anchors and positive samples) closer to each other in the representation space, while pushing them away from the vectors representing unfaithful answers (negative samples). To this end, we employ the InfoNCE (Noise-Contrastive Estimation) loss function[22], defined as:

$$L_{contrastive} = -\log \frac{\exp(\text{sim}(h_{\text{anchor}}, h_{\text{pos}})/\tau)}{\exp(\text{sim}(h_{\text{anchor}}, h_{\text{pos}})/\tau) + \sum_{i=1}^{N} \exp(\text{sim}(h_{\text{anchor}}, h_{\text{neg}_i})/\tau)} \quad (2)$$

where:

(1) $h_{anchor}$, $h_{pos}$, $h_{neg_i}$ are the representation vectors of the anchor, positive sample, and the $i$-th negative sample.

(2) $N$ is the number of negative samples per anchor (in this study, $N = 3$).

(3) $sim(·,·)$ is the cosine similarity function, measuring the directional consistency between two vectors.

(4) $\tau$ is a temperature hyperparameter used to adjust the distribution of similarity scores. A lower temperature amplifies the differences between similar samples, making the model focus more on distinguishing hard negative samples.

In this way, the model learns a representation space where contextual faithfulness becomes a significant, separable semantic feature. This fundamentally enhances the model's reliance on context during generation and suppresses its tendency to rely on internal conflicting knowledge, thereby successfully completing the learning phase of the self-improvement loop. Algorithm 1 shows the training procedure for the SI-FACT contrastive loss.

---

Algorithm 1: Contrastive Loss Calculation

**Input:** LLM, $(C, Q, A_{anchor})$ (Anchor triplet), $A_{pos}$ (Positive sample), $A_{neg\_set}$ (Set of negative samples), $\tau$ (Temperature)

**Output:** $L_{contrastive}$

1: $h_{anchor} \leftarrow$ get_representation(LLM, concat($C, Q, A_{anchor}$)) // Extract Representations

2: $h_{pos} \leftarrow$ get_representation(LLM, concat($C, Q, A_{pos}$))

3: $positive\_score \leftarrow$ cosine_similarity($h_{anchor}$, $h_{pos}$)// Calculate Similarity & Scores

4: $numerator \leftarrow exp(positive\_score / \tau)$

5: $denominator \leftarrow numerator$ // Initialize denominator with the positive pair's score

6: for each $A_{neg}$ in $A_{neg\_set}$ do

7: $\quad h_{neg} \leftarrow$ get_representation(LLM, concat($C, Q, A_{neg}$))

8: $\quad negative\_score \leftarrow$ cosine_similarity($h_{anchor}$, $h_{neg}$)

9: $\quad denominator \leftarrow denominator + exp(negative\_score / \tau)$

10: end for

11: $L_{contrastive} \leftarrow -log(numerator / denominator)$

12: return $L_{contrastive}$

---

## 3. Experiments and Analysis

This section shows a series of experiments conducted to validate the effectiveness of the SI-FACT framework, including the experimental setup, main results, and in-depth analysis.

### 3.1 Experimental Setup

#### 3.1.1 Evaluation Datasets and Metrics

**Datasets**: Experiments were conducted on two benchmark datasets widely used to evaluate LLM performance in knowledge conflict scenarios: ECARE_KRE and COSE_KRE. These datasets contain machine-generated misleading contexts that contradict the model's internal parametric knowledge (i.e., common sense). This makes them ideal datasets for examining whether the model will adhere to its pre-trained knowledge or prioritize the current context.

**Evaluation Metrics**: To comprehensively evaluate model behavior, we defined the following three metrics:

(1) Contextual Recall Rate (CRR): Measures the proportion of model-generated answers that are consistent with the given context. A higher CRR indicates a stronger ability to follow the context:

$$CRR = \frac{Contextual\ answers\ count}{Total\ questions} \times 100\% \quad (3)$$

(2) Parametric Recall Rate (PRR): Measures the proportion of answers that are consistent with its internal parametric knowledge. This metric reflects the model's stubbornness. A lower PRR indicates a greater emphasis on the context:

$$PRR = \frac{Parametric\ answers\ count}{Total\ questions} \times 100\% \quad (4)$$

(3) Memorization Ratio (MR): Comprehensively evaluates the trade-off between the model's reliance on context versus internal memory. A lower MR value indicates that the model relies much more on context than on internal memory. It is calculated as:

$$MR = \frac{PRR}{CRR + PRR} \quad (5)$$

#### 3.1.2 Baseline

We compare SI-FACT with a series of representative baseline methods, covering a range of techniques from simple to complex:

**Original**: The base Llama3-8B-Instruct[23] model without any fine-tuning. **Context_Prompt**: Add the simple prompt "Based on the context" before the question; **Formatted Context_Prompt**: Format the input as "Context: ... Question: ..."; **Enhanced Context_Prompt**: Use the more explicit prompt "Using the knowledge in the context" and employ formatted input. **Opin**[7]: An advanced prompt engineering method that transforms factual questions into opinion-based ones to encourage reliance on the context. Inference-time intervention methods **CAD**[18] and **IRCAN**[8].

3.1.3 Implementation Details

All experiments were conducted based on the Llama3-8B-Instruct model. Within the SI-FACT framework, we used the model to build a total of 12000 contrastive learning data instances from the SQuAD dataset via the aforementioned self-instruct data generation process, each containing an anchor, a positive sample, and three types of negative samples. A key finding was that the model's performance tended to saturate and reach its peak with only 1000 training samples. This finding highlights the high data efficiency of the SI-FACT method.

## 3.2 Results and Analysis

3.2.1 Analysis of the performance comparison with the baseline

We comprehensively compared the performance of SI-FACT against all baseline methods on the COSE_KRE and ECARE_KRE datasets. Detailed results are shown in Table 2 and Table 3, with the best value in bold and the second best underlined.

Table 2  Performance Comparison on the ECARE_KRE Dataset

| Datasets | Methods | CRR (↑) | PRR (↓) | MR (↓) |
|---|---|---|---|---|
| **ECARE_KRE** | Original | 57.30 | 42.70 | 0.427 |
| | Context_Prompt | 59.10 | 40.90 | 0.409 |
| | Formatted Context_Prompt | 68.90 | 31.10 | 0.311 |
| | Enhanced Context_Prompt | 69.18 | 30.82 | 0.308 |
| | Opin | 68.05 | 31.95 | 0.320 |
| | CAD | 69.75 | 30.25 | 0.303 |
| | IRCAN | 57.87 | 42.13 | 0.421 |
| | SI-FACT (Ours) | 75.97 | 24.03 | 0.240 |

The results clearly demonstrate the superior performance of the SI-FACT framework. On the core metric CRR, SI-FACT achieved the highest scores on both datasets. On the ECARE_KRE dataset, SI-FACT's advantage was most pronounced, with a CRR of 75.97%, exceeding that of CAD[18] (69.75%) by over 6.2%. On the more challenging COSE_KRE dataset, SI-FACT's CRR reached 54.17%, also surpassing the next best method, CAD[18] (52.86%). This indicates a fundamental enhancement in the model's ability to adhere to the given context.

On the PRR metric, which evaluates the stubbornness, SI-FACT also performed best, achieving the lowest values on both datasets. On ECARE_KRE, SI-FACT's PRR was only 24.03%, significantly lower than all baseline methods. This proves that SI-FACT not only improves faithfulness but also most effectively suppresses the model's tendency to rely on its internal conflicting knowledge.

Table 3  Performance Comparison on the COSE_KRE Dataset

| Datasets | Methods | CRR (↑) | PRR (↓) | MR (↓) |
|---|---|---|---|---|
| **COSE_KRE** | Original | 39.93 | 47.63 | 0.544 |
| | Context_Prompt | 42.39 | 45.50 | 0.518 |
| | Formatted Context_Prompt | 51.72 | 38.79 | 0.429 |
| | Enhanced Context_Prompt | 50.08 | 40.10 | 0.445 |
| | Opin | 52.54 | 36.17 | 0.407 |
| | CAD | 52.86 | 35.84 | 0.404 |
| | IRCAN | 42.72 | 37.64 | 0.468 |
| | SI-FACT (Ours) | 54.17 | 33.22 | 0.380 |

Finally, the comprehensive metric MR (Memorization Ratio) further confirms the superiority of SI-FACT. SI-FACT achieved the lowest MR values on both datasets. This shows that it achieves the best balance between enhancing context adherence (high CRR) and suppressing memory dependence (low PRR), representing a comprehensive improvement.

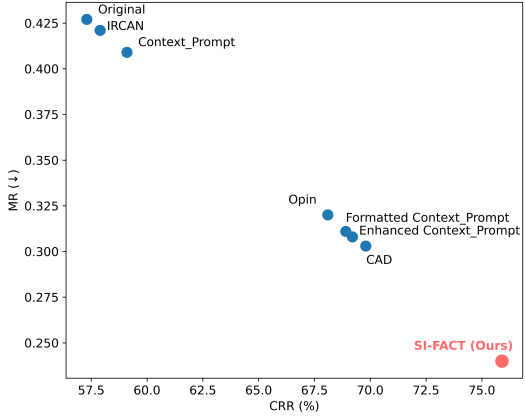

Fig. 2  MR–CRR Frontier Scatter Plot

lower MR). It can be seen that SI-FACT is located at the lower-right corner of the performance frontier, significantly outperforming other methods.

### 3.2.2 Representation Space Analysis

To visually verify the effect of SI-FACT in "pulling faithful representations closer and pushing unfaithful ones farther apart" in the representation space, we performed centralized t-SNE[24] visualization on the vectors of the Anchor, Positive, and Negative samples for multiple data points. Specifically, for each data point, we first extracted the sentence embedding vectors $h$ for the Anchor and its corresponding positive and negative samples. We then centralized them using the Anchor's representation vector as the origin ($\Delta h = h - h_{anchor}$) to eliminate semantic shifts between different content. Subsequently, all centralized difference vectors were aggregated, and t-SNE was used to reduce the high-dimensional embeddings to two dimensions.

To more intuitively compare the comprehensive performance of different methods, we plotted an MR-CRR frontier scatter plot (Fig. 2). The x-axis is CRR, and the y-axis is MR. Points located towards the bottom-right indicate better performance (higher CRR and

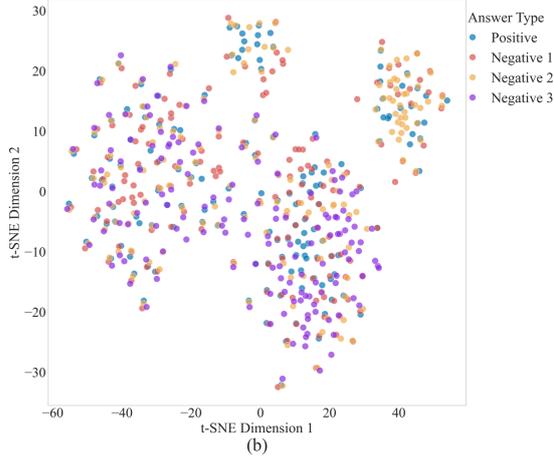
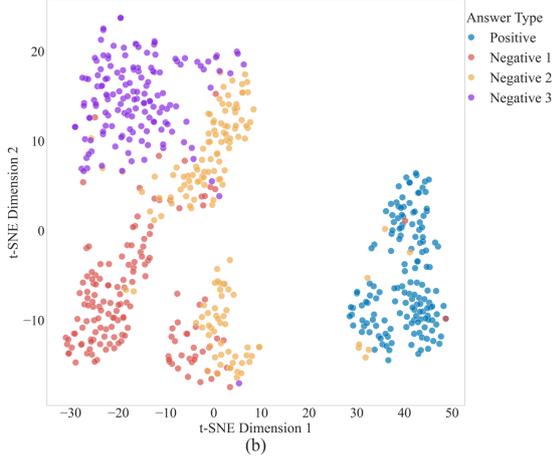

Fig. 4  Centralized t-SNE visualization results. (a) Before training; (b) After SI-FACT training

Fig. 4(a) shows the results before training, where positive and negative samples are intermingled in the space without clear boundaries, indicating that the model did not distinguish between context-faithful and unfaithful answers in the representation space before SI-FACT optimization.

Fig. 4(b) displays the results after SI-FACT training. All positive samples, representing context-faithful answers, are clearly clustered to the right of the origin in the 2D projection space, indicating they are closer to their respective Anchors. The negative samples are generally distributed on the other side and are significantly more separated from the positive samples. This demonstrates that SI-FACT has encoded faithfulness as a separable direction, making the distinction between context-faithful and unfaithful answers more pronounced at the representation level.

It is worth noting that the model not only learned to distinguish between context-faithful and unfaithful answers but also, to some extent, formed distinct sub-clusters for different types of unfaithful answers.

### 3.2.3 Data Efficiency Analysis

A core finding of this study is the extremely high data efficiency of the SI-FACT framework. Experiments showed that performance peaked after using only 1000 self-generated training samples.

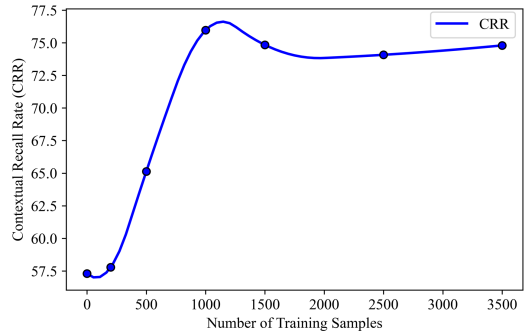

Fig. 5 Performance curve of the SI-FACT framework under different training sample sizes

This result demonstrates the high efficiency of SI-FACT in improving the contextual faithfulness of LLM. The targeted data actively generated by the model is far more efficient than random or large-scale but less targeted data. This high efficiency makes SI-FACT a highly practical and scalable solution, enabling the enhancement of LLM reliability under resource constraints and overcoming the major obstacle that pre-hoc methods are often considered costly.

3.2.4 Analysis of Impact on General Capabilities

To verify whether our proposed SI-FACT framework would harm the model's existing general capabilities, we conducted a comparison of the performance of the SI-FACT tuned model and the original model on four widely recognized general capability benchmarks. These benchmarks cover knowledge question answering (TriviaQA)[25], mathematical reasoning (GSM8K)[26], commonsense reasoning (Hellaswag)[27], and complex question answering (ARC-Challenge)[28].

Table 4  Performance Comparison on General Capability Benchmarks

| Benchmark | Task Type | Original Model Acc. (%) | SI-FACT Model Acc. (%) | Performance Change |
| --- | --- | --- | --- | --- |
| TriviaQA | Knowledge QA | 71.00 | 71.40 | +0.40 |
| GSM8K | Mathematical Reasoning | 67.40 | 66.60 | -0.80 |
| Hellaswag | Commonsense Reasoning | 65.40 | 64.71 | -0.75 |
| ARC-Challenge | Complex QA | 74.41 | 72.48 | -1.93 |

As shown in Table 4, the SI-FACT tuned model exhibited a slight performance decrease on some general benchmarks but showed a minor improvement on the knowledge Q&A benchmark, TriviaQA. This result indicates that our method largely preserves the model's core general reasoning and understanding abilities. This minor performance trade-off is acceptable, suggesting that SI-FACT is an efficient and targeted enhancement method. Its optimization objective is focused on improving contextual faithfulness without significantly damaging the model's existing knowledge system. This further highlights the practical value of our framework as a minimally invasive optimization strategy.

## 4. Conclusions

This paper proposed the SI-FACT framework to address the issue of Large Language Models generating unfaithful answers in knowledge conflict scenarios. The framework uses a self-instruction mechanism to generate high-quality contrastive samples and leverages contrastive learning to successfully strengthen the model's capability for contextual faithfulness at the representation level. Experimental results show that SI-FACT not only significantly outperforms existing methods on the key metrics of CRR, PRR, and MR, but also achieves optimal performance with only a small amount of data, highlighting its efficiency. Furthermore, SI-FACT did not significantly affect the model's general capabilities, indicating that it achieves targeted optimization while preserving its original reasoning and understanding abilities.

From a theoretical perspective, this method provides empirical evidence for the feasibility of model self-teaching, enriching the application scenarios of self-supervised and contrastive learning in the optimization of large models. From a practical standpoint, SI-FACT can enhance a model's contextual faithfulness at a low annotation cost, offering a viable path for deploying trustworthy LLMs in high-risk applications such as finance and healthcare.

Future research will extend SI-FACT in three directions: applying it to larger-scale models; integrating it with RAG to optimize faithfulness in complex, open environments; and extending the self-improvement paradigm to other critical capabilities such as safety enhancement and bias mitigation, thereby advancing LLMs toward greater autonomy and reliability.